\documentclass[final]{cvpr}

\usepackage{times}
\usepackage{epsfig}
\usepackage{graphicx}
\usepackage{amsmath}
\usepackage{amssymb}

\usepackage{fdsymbol}
\usepackage{multirow}
\usepackage{bbm}

\DeclareMathOperator*{\argmin}{arg\,min}
\newcommand{\distas}[1]{\mathbin{\overset{#1}{\kern\z@\sim}}}%

\usepackage[ruled,vlined]{algorithm2e}

\SetCommentSty{mycommfont}
\SetKwInput{KwInput}{Input}                
\SetKwInput{KwOutput}{Output}              

\usepackage{booktabs}
\newcommand{\bftab}{\fontseries{b}\selectfont}

\usepackage[pagebackref=true,breaklinks=true,colorlinks,bookmarks=false]{hyperref}

\begin{document}

\title{Hybrid Consistency Training with Prototype Adaptation\\ for Few-Shot Learning}

\author{Meng Ye\quad Xiao Lin\quad Giedrius Burachas\quad Ajay Divakaran\quad Yi Yao\\
SRI International\\
{\tt\small \{meng.ye, xiao.lin, giedrius.burachas, ajay.divakaran, yi.yao\}@sri.com}
}

\maketitle

\begin{abstract}
Few-Shot Learning (FSL) aims to improve a model's generalization capability in low data regimes. Recent FSL works have made steady progress via metric learning, meta learning, representation learning, etc. However, FSL remains challenging due to the following longstanding difficulties. 1) The seen and unseen classes are disjoint, resulting in a distribution shift between training and testing.
2) During testing, labeled data of previously unseen classes is sparse,
making it difficult to reliably extrapolate from labeled support examples to unlabeled query examples. To tackle the first challenge, we introduce Hybrid Consistency Training to jointly leverage  interpolation consistency, including interpolating hidden features, that imposes linear behavior locally and data augmentation consistency
that learns robust embeddings against sample variations. 
As for the second challenge, we use unlabeled examples to iteratively normalize features and adapt prototypes, as opposed to commonly used one-time update, for more reliable prototype-based transductive inference.
We show that our method generates a $2\%$ to $5\%$ improvement over the state-of-the-art methods with similar backbones on five FSL datasets and, more notably, a $7\%$ to $8\%$ improvement for more challenging cross-domain FSL. 
\end{abstract}

\section{Introduction}
\begin{figure}[t]
\begin{center}
   \includegraphics[width=0.95\linewidth]{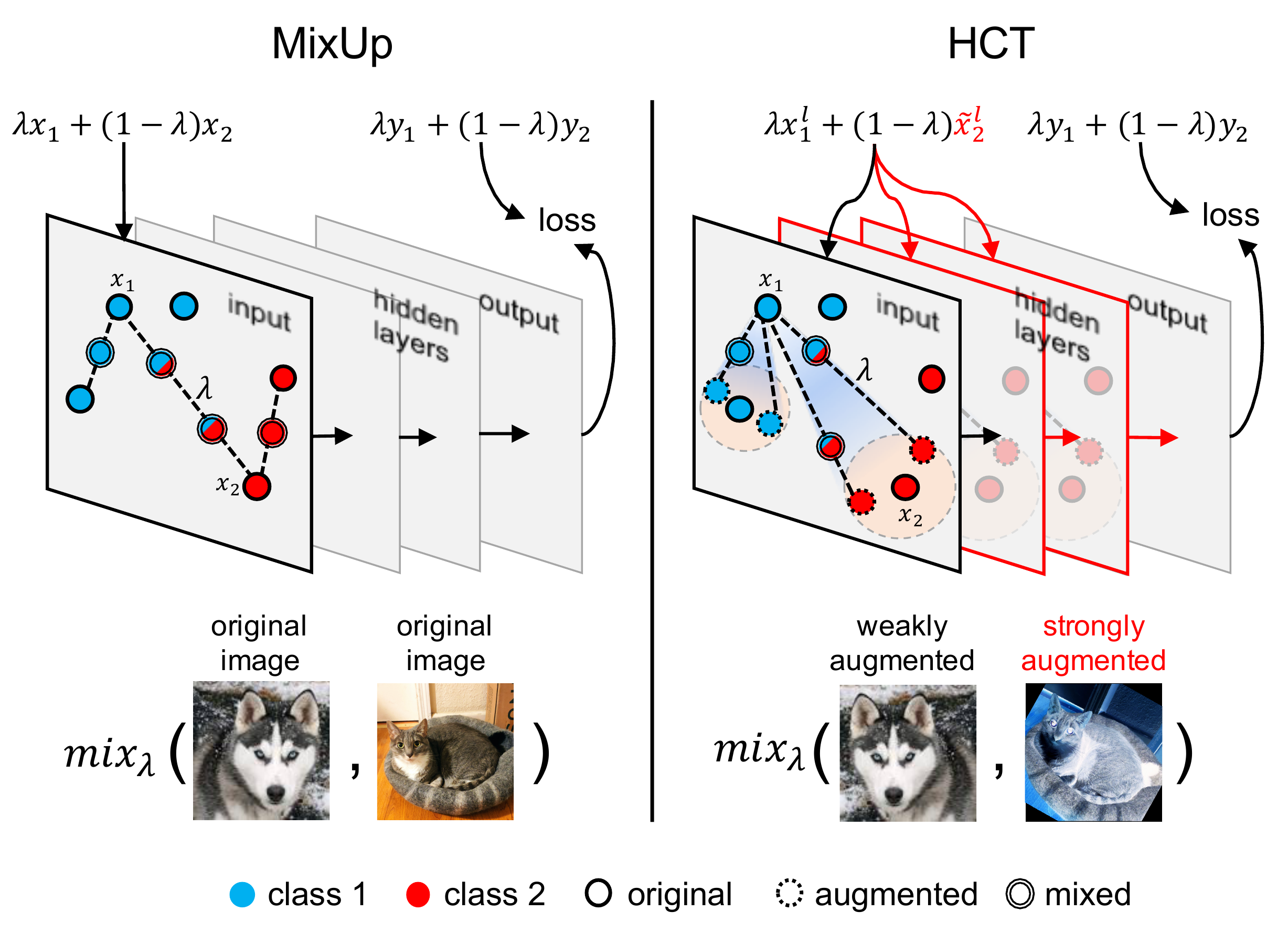}
\end{center}
   \caption{Comparison between Mixup~\cite{zhang2017mixup} and Hybrid Consistency Training (HCT). Mixup imposes interpolations on lines between two examples.
   In HCT, the strongly augmented image is further away from the original image. Thus, interpolations cover a wider range in the input/feature space,
   resulting in a stronger regularization for FSL (best viewed in color).}
\vspace{-1em}
\label{fig:intro}
\end{figure}
Despite its successful applications in various computer vision tasks,
deep learning still remains challenging in low data regimes.
Recently, Few-Shot Learning (FSL) has draw increasing attention in various computer vision tasks, including 
image classification~\cite{hariharan2017low,ravi2016optimization,snell2017prototypical,sung2018learning,vinyals2016matching},
object detection~\cite{kang2019few,karlinsky2019repmet} and semantic segmentation~\cite{siam2019amp,dong2018few}.
In FSL, the training classes (\ie, seen or \emph{base} classes) and the testing classes (\ie, unseen or \emph{novel} classes) are disjoint.
In order to perform classification on \emph{novel} classes using only a few labels,
certain form of knowledge must be learned and transferred from \emph{base} to \emph{novel} classes.
Such knowledge can be a metric space~\cite{snell2017prototypical,vinyals2016matching,koch2015siamese},
a model initialization~\cite{finn2017model}, a learning algorithm~\cite{ravi2016optimization} , or
simply an embedding model~\cite{chen2020new,tian2020rethink}.
While having demonstrated success on few-shot tasks, these approaches still fall short in addressing the following longstanding challenges: 1) large semantic gap between \emph{base} and \emph{novel} classes and 2) sparsity of labeled data of \emph{novel} classes. 

To tackle semantic gaps between \emph{base} and \emph{novel} classes, learning richer features to reduce overfitting on the \emph{base} classes via incorporating  knowledge learned from the images themselves is a promising direction  \cite{tian2020rethink}. For example, self-supervised losses, such as rotation~\cite{gidaris2019boosting} and exemplars~\cite{mangla2020charting}, are employed in addition to the supervised loss on \emph{base} classes
for improved features
~\cite{doersch2015unsupervised,noroozi2016unsupervised,gidaris2018unsupervised}. 
In stead of constructing explicit surrogate tasks, another popular line of works exploit additional regularization such as consistency losses, inspired by semi-supervised learning. 
For example, interpolation consistency~\cite{Verma2019ict,verma2019manifold,zhang2017mixup}
encourages a model's local linearity
and data augmentation consistency~\cite{xie2019unsupervised,berthelot2019remixmatch,sohn2020fixmatch} enforces a model's local continuity.

In this paper, we propose Hybrid Consistency Training (HCT), which uniquely combines the above two consistencies by directly imposing interpolation linearity on top of weakly and strongly augmented samples across intermediate features, as opposed to commonly used post-hoc combination of two independent losses (Fig.~\ref{fig:intro}). 
Specifically, we construct mixed features at a randomly selected network layer using a weakly and strongly augmented samples from a pair of labeled input images.  
The loss is measured by the cross entropy between model predictions of such mixed features and the linear combination of the ground truth labels of the original input images.
Intuitively, weakly and strongly augmented samples reside in a
smaller (with limited variations) and a larger (with richer variations)
neighborhood of the original image, respectively. Applying interpolation consistency on strongly augmented samples enforces local continuity and linearity in a wider range, leading to richer yet more regularized embedding space. Moreover, applying interpolation consistency across intermediate features further smoothens decision boundaries throughout all network layers. Richer yet flattened (\ie, with fewer directions of variance) representations and smoother decision boundaries lead to improved generalization capability despite large semantic gaps. 


The second challenge stems from the sparsity of labeled samples from \emph{novel} classes. In this regard, transductive inference is introduced to leverage unlabeled data to fill in the gaps between labeled and query examples \cite{liu2018learning}. In this work, we advance prototype-based transductive inference by introducing an iterative method
to calibrate features and 
adapt prototypes of \emph{novel} classes using unlabeled data, referred to as Calibrated Iterative Prototype Adaptation (CIPA). While being simple, feature calibration (\eg, power transformation, centering, normalization) is a critical step that aligns samples from the support and query/unlabeled sets, producing an improved common ground for distance computation.
Meanwhile, by estimating pseudo-labels on unlabeled data and updating prototypes iteratively, prototype estimations can be more precise despite the sparse and non-uniformly distributed labeled samples. Compared to another iterative method~\cite{hu2020leveraging}, where Sinkhorn~\cite{cuturi2013sinkhorn} mapping is employed for pseudo labeling unlabeled data, our CIPA uses simple but effective cosine similarity, which requires much less computation. More critically, \cite{hu2020leveraging} assumes equal number of examples per class. In contrast, our CIPA does not rely on such assumptions and can work properly even under class imbalance.

Our contributions are:

1) We propose a Hybrid Consistency Training method built upon both interpolation and data augmentation consistencies to enforce local linearity and continuity in a wider extent (\ie, by incorporating strongly augmented samples) and across all network layers (\ie, by using Manifold Mixup). This generates significantly stronger embeddings to support generalization across large semantic gaps between the \emph{base} and \emph{novel} classes for improved FSL.

2) We propose an iterative prototype-based transductive inference algorithm to calibrate features and adapt class prototypes using unlabeled data.
This can leverage unlabeled data to effectively fill the gaps between query and labeled samples, which are sparse and frequently non-uniformly distributed.

3) Through extensive experiments we show that our method generates a $2\%$ to $5\%$ improvement over the state-of-the-art (SOTA) methods with similar backbones on five FSL datasets and, more notably, a $7\%$ to $8\%$ improvement for more challenging cross-domain FSL (\eg, \emph{mini}-ImageNet to CUB). 


\section{Related work}

\subsection{Few-shot learning}

\emph{Metric learning} methods learn a metric function from the base classes and use it to measure distance for novel data.
Some prior work uses learnable parameters to model the metric function, for example a linear layer~\cite{koch2015siamese}, LSTM~\cite{vinyals2016matching}
or convolutional networks~\cite{sung2018learning}. Others learn a backbone network as embedding functions and use fixed metric
to compute classification scores, such as euclidean distance~\cite{snell2017prototypical}, cosine similarity~\cite{gidaris2018dynamic,chen2019closer} and Mahalanobis distance~\cite{bateni2020improved}. More recently, researchers started looking closer into image regions for calibrated metric spaces, \eg, \cite{zhang2020deepemd} finds correspondences between two sets of image regions using earth mover's distance
and \cite{hou2019cross} proposes a cross-attention network to focus on representative image regions.

Instead of learning a metric function, \emph{optimization-based meta-learning} methods extract meta-knowledge from the training classes and apply it on novel data.
MAML~\cite{finn2017model} learns a good model initialization that can reach optimum with a few steps of gradient descent.
\cite{ravi2016optimization} uses LSTM as a meta-learner to learn the optimization algorithm itself that can reach convergence fast on novel classes.
LEO~\cite{rusu2018meta} performs meta-learning using a low-dimensional space for model parameter representations.

Despite the progress in meta-learning, some recent work shows that by training a representation model on all the base classes,
the resulting embeddings can be quite effective for FSL. We refer to these as \emph{representation learning} based methods.
In \cite{chen2020new}, it is shown that using distances computed on pre-trained embedding using base classes already achieve competitive results.
\cite{tian2020rethink} shows that learning a supervised representation from base classes followed by 
training a linear classifier on those representations for novel classes can also be quite effective.
Tian \etal~\cite{tian2020rethink} report similar observations. 
Compared to complex meta-learning approaches, \emph{representation learning} based methods are much simpler and still effective in generalizing 
knowledge learned from base to novel data.

Besides the methods mentioned above, another line of work incorporates self-supervised learning~\cite{dosovitskiy2014discriminative,doersch2015unsupervised,noroozi2016unsupervised,gidaris2018unsupervised,caron2018deep,chen2020simple} for FSL.
For example, \cite{gidaris2019boosting} finds that adding a rotation prediction task alongside the classification task
to train a network leads to better FSL performance.
Su~\etal~\cite{Su2020When} note that self-supervised learning can bring greater improvements on smaller datasets or more challenging tasks.
Mangla~\etal~\cite{mangla2020charting} use Manifold Mixup~\cite{verma2019manifold} regularization as well as 
self-supervision loss terms (rotation~\cite{gidaris2018unsupervised} and exemplar~\cite{dosovitskiy2014discriminative}) to learn robust representations.

Our HCT method 
is based on representation learning. It is orthogonal to self-supervised techniques and can be combined with them by adding more losses in a multi-task learning manner.

\subsection{Semi-supervised learning with consistency}
\label{sec:consistency}
Semi-supervised learning aims at leveraging unlabeled data in addition to labeled data
to perform given tasks.
Here we discuss a few semi-supervised methods using consistency-based regularization, which is closely related to our work.
Virtual Adversarial Training (VAT)~\cite{miyato2018virtual} finds local adversarial perturbations and enforces consistent model predictions despite such perturbations.
FixMatch~\cite{sohn2020fixmatch} is a combination of pseudo-labeling and data augmentation-based consistency regularization.
For an unlabeled image, a weakly and a strongly augmented versions are generated.
The weak version is used to obtain the pseudo-label for the strong version.
Interpolation Consistency Training (ICT)~\cite{Verma2019ict} extends Mixup~\cite{zhang2017mixup} to unlabeled data for semi-supervised learning.
It uses interpolation consistency: given an interpolation of two examples as input, 
the model should be consistent to output the interpolation of their predictions.
These consistency constrains regularize network training so that the learned networks can generalize better on test data.

Borrowing ideas from semi-supervised learning,
our HCT combines interpolation and data augmentation consistency and applies these consistency-based losses on labeled data from base classes.
By generating an interpolation between a weakly and a strongly augmented examples,
we enforce the model output to be consistent with respect to the interpolation of their labels.
We also regularize network training by not just interpolating the input images,
but also interpolating the hidden features.
In so doing, we introduce stronger regularization and, therefore, expect smoother manifolds.

\section{Method}

\begin{figure*}[ht]
\begin{center}
   \includegraphics[width=0.9\linewidth]{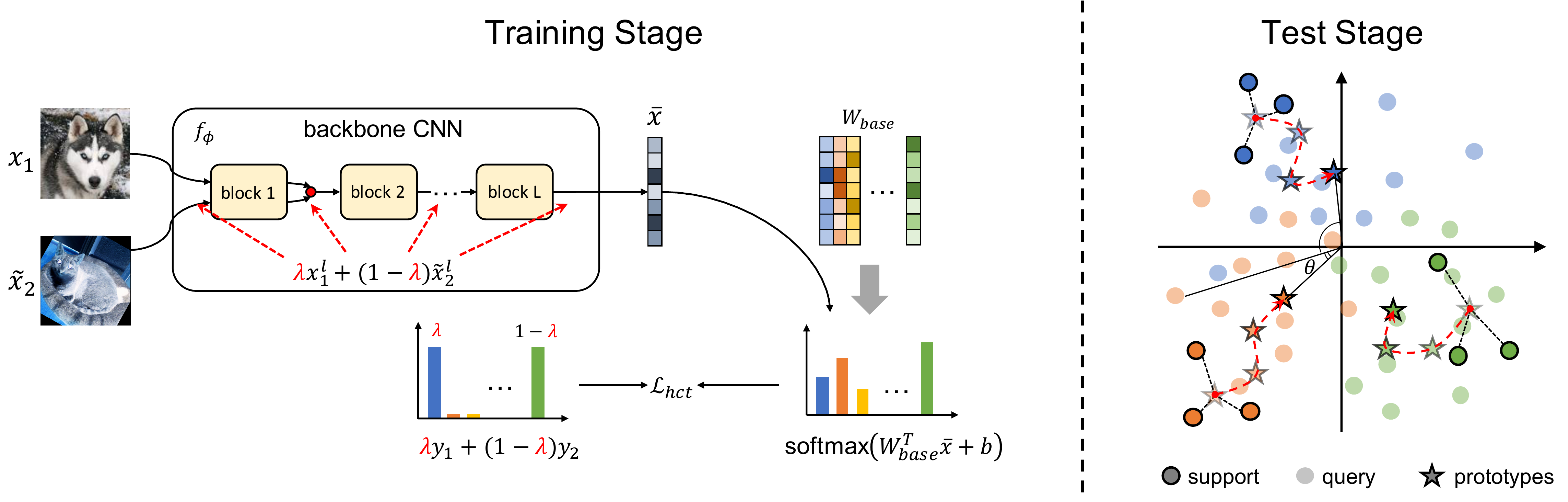}
\end{center}
\vspace{-1em}
\caption{An overview of our proposed approach.
Left: Hybrid Consistency Training (HCT) for embedding learning. In the training stage, an embedding network is learned on the classification task over all base classes.
We add the hybrid consistency loss to regularize the training for smoother hidden feature manifolds.
Right: Calibrated Iterative Prototype Adaptation (CIPA) for transductive inference. In the testing stage, we transform the features of each task to calibrate its data distribution.
Then we apply an iterative algorithm to adapt unlabeled data and produce better class prototype estimations (best viewed in color).}
\vspace{-1em}
\label{fig:overview}
\end{figure*}

In FSL, it is commonly assumed that there is a training dataset $\mathcal{D}_{base}$
of base classes $\mathcal{C}_{base}$ and a test dataset $\mathcal{D}_{novel}$ of novel classes $\mathcal{C}_{novel}$.
These two sets of classes are totally disjoint $\mathcal{C}_{base}\cap\mathcal{C}_{novel}=\varnothing$.
Depending on different FSL approaches, the base dataset can be used either as a single dataset $\mathcal{D}_{base}=\{x_i, y_i\}_{i=1}^{N_{base}}$ ($x$ and $y$ denote image and label, respectively),
or a source for sampling few-shot tasks (or episodes) 
$\mathcal{T}_{base}=\{(\mathcal{D}^S_i, \mathcal{D}^Q_i)\}_{i=1}^{N_{episode}}$, where
$\mathcal{D}^S_i=\{(x^s_i,y^s_i)\}_{i=1}^{NK}$ is the support set with $NK$ labeled examples
and $\mathcal{D}^Q_i=\{(x^q_i,y^q_i)\}_{i=1}^{NQ}$ is the set with $NQ$ query examples.
This is typically referred to as an $N$-way $K$-shot problem.
For evaluation, a number of novel tasks are sampled from the test dataset
$\mathcal{T}_{novel}=\{(\mathcal{D}^S_i, \mathcal{D}^Q_i)\}_{i=1}^{N_{episode}}$
similarly in $N$-way $K$-shot episodes
and the average accuracy on these episodes is used as the final measure of performance.

\subsection{Background}
Given the  base and novel datasets, our goal is to learn an embedding network $f_\phi$
from the \emph{base} data, so that it can be used to compute distances between \emph{novel} images for prediction.

\noindent{\bf Prototypical Network}~\cite{snell2017prototypical} uses the centroid of support examples
from each class $c$ as its prototype.
The distances between a query example and all the prototypes
are computed, and then a softmax operation is applied to output the class distribution:
\begin{align}
    \mathbf{p}_c &=\frac{1}{K}\sum_{x^s_i\in\mathcal{D}^S}\mathbbm{1}_{[y^s_i=c]}f_\phi(x^s_i) \label{eq:proto}\\
    p_c(x^q) &= \frac{\mathrm{exp}\Big(-\tau\cdot d\big(\mathbf{p}_c,f_\phi\left(x^q\right)\big)\Big)}
                    {\sum_{c'}\mathrm{exp}\Big(-\tau\cdot d\big(\mathbf{p}_{c'},f_\phi\left(x^q\right)\big)\Big)}, \label{eq:softmax_proto}
\end{align}
where $d(\cdot)$ is the metric function, \eg, euclidean distance or negative cosine similarity and $\tau$ is a scalar.
The network is trained by minimizing a loss function defined as the cross-entropy of each query instance for all training episodes:
\begin{align}
    \phi^*&=\argmin_{\phi}\mathop{\mathbb{E}}_{(\mathcal{D}^S,\mathcal{D}^Q)\in\mathcal{T}_{base}}\mathcal{L}_{ce}(\mathcal{D}^S,\mathcal{D}^Q) \nonumber\\
        &=\argmin_{\phi}\mathop{\mathbb{E}}_{(\mathcal{D}^S,\mathcal{D}^Q)\in\mathcal{T}_{base}}\sum_{(x^q_i,y^q_i)\in\mathcal{D}^Q}-\mathrm{log}\,p_{y^q_i}(x^q_i)
\end{align}
The purpose of episodic training is to simulate the few-shot evaluation protocol and reduce over-fitting on the base classes.

\vspace{1mm}
\noindent{\bf Classifier Baseline} is a simple FSL method that learns an embedding from all base data.
Just as in standard supervised learning, a fully connected layer is appended on top of $f_\phi$ to output logits for each base class.
By sampling batches of images from $\mathcal{D}_{base}$, the embedding network can be learned by minimizing the cross entropy loss on 
model output and the ground truth labels:
\begin{equation}
    p(x) = \mathsf{softmax}\big(W_{base}^Tf_\phi\left(x\right)+b\big)
\end{equation}
and
\begin{align}
    \phi^*  &= \argmin_{\phi}\mathop{\mathbbm{E}}_{(x_i,y_i)\in\mathcal{D}_{base}}\mathcal{L}_{ce}(x_i,y_i) \nonumber\\
            &= \argmin_{\phi}\mathop{\mathbbm{E}}_{(x_i,y_i)\in\mathcal{D}_{base}}\sum_{c=1}^{|\mathcal{C}_{base}|}-y_{i,c}\,\mathrm{log}\,p_c(x_i) \label{eq:ce}
\end{align}

The above two methods lay a good foundation, upon which various techniques can be added to improve FSL performance.

\subsection{Hybrid Consistency Training}
In this section, we introduce HCT,
which can be viewed as a regularization technique that improves \emph{representation learning} for few-shot task (Fig.~\ref{fig:overview} left panel).
Assume that the embedding function is a composition of multiple layers
$f_\phi=f^L\circ\cdots\circ f^1\circ f^0$.
The hidden representation at layer $l$ can be obtained by passing the input image through layer $0,1\cdots l$: $h^l=f^l\circ\cdots\circ f^1\circ f^0(x)$. Note that $f^0$ is the input layer and $h^0=f^0(x)=x$.
Given an embedding model, we optimize its weights by minimizing the following loss function
\begin{equation}
    \mathcal{L}=\mathcal{L}_{ce}+\eta\mathcal{L}_{hct},
\end{equation}
where $\mathcal{L}_{ce}$ is the cross entropy loss on the base classes as in Eq.~(\ref{eq:ce}),
$\eta$ is a balancing parameter (we set it to $1$ in all our experiments), and
$\mathcal{L}_{hct}$ is our newly introduced hybrid consistency loss 
which we explain in details below.

As mentioned in Sec.~\ref{sec:consistency}, consistency training has been widely used in semi-supervised learning.
In this work, we propose HCT, combining two different consistency training approaches
into a unified framework to regularize model training.
Given any two images $x_1$ and $x_2$, we perform weak augmentation, \eg horizontal flip, to $x_1$
so that the augmented image is still close to the original image. We overload the notation $x_1$ to represent both the original image and its weakly augmented version.
For $x_2$, we apply strong augmentation (see details in Sec.~\ref{sec:settings}) so that it is heavily distorted and has a higher chance of
being out of the local data distribution. We denote this example as $\tilde{x}_2$.
To generate an interpolation between $x_1$ and $\tilde{x}_2$,
we feed them both into the embedding network and then randomly choose a layer $l$ to get their hidden representations:
\begin{align}
    x^l_1 &= f^l\circ\cdots f^1\circ f^0(x_1) \nonumber \\ 
    \tilde{x}^l_2 &= f^l\circ\cdots f^1\circ f^0(\tilde{x}_2).
\end{align}
The hidden representations are mixed and passed through remaining layers
to get the final feature representation $\bar{x}$:
\begin{align}
    \bar{x}^l &= \lambda\cdot x^l_1 + (1-\lambda)\cdot\tilde{x}^l_2 \nonumber\\
    \bar{x} &= f^L\circ\cdots f^{l+1}(\bar{x}^l).
\end{align}
The corresponding target $\bar{y}$ is the interpolation of the ground truth one-hot label vectors $y_1$ and $y_2$ of the original input samples $x_1$ and $x_2$:
\begin{align}
    \bar{y} = \lambda\cdot y_1 + (1-\lambda)\cdot y_2.
\end{align}
Then, the loss function on these interpolated examples is
\begin{align}
    \mathcal{L}_{hct} = &\mathop{\mathbbm{E}}_{
                            \substack{(x_1,y_1)\in\mathcal{D}_{base}\\(x_2,y_2)\in\mathcal{D}_{base}\\
                            \lambda\sim Beta(\alpha,\alpha),\ l\sim U(0,L)}}
                        \sum_{c=1}^{|\mathcal{C}_{base}|}-\bar{y}_{c}\,\mathrm{log}\,p_c(\bar{x}).
\end{align}

HCT combines interpolation consistency~\cite{zhang2017mixup,Verma2019ict}
and data augmentation consistency~\cite{xie2019unsupervised,berthelot2019remixmatch,sohn2020fixmatch}
in a unique and tightly integrated way: the generated new data points not only cover linear space between examples,
but also expand further to the regions where heavily distorted examples reside.
By doing this at a random layer each time, hidden representations at all levels are regularized.
This leads to a smoother manifold that generalizes better to novel classes.
HCT can also be combined with other representation learning techniques, \eg, self-supervised rotation classification $\mathcal{L}_{rot}$, by simply adding another head and performing multi-task learning (denoted as HCT\textsubscript{R}),
which often results in further improved representations.

\subsection{Calibrated Iterative Prototype Adaptation}
We use the embedding model $f_{\phi^*}$ trained using HCT described in the previous section to infer predictions for novel data.
Given a novel task $\mathcal{T}_{novel}^{(i)}=(\mathcal{D}^S_i,\mathcal{D}^Q_i)$,
we first extract features of both the support examples and the query examples.
A straightforward way to get class probabilities is to compute class prototypes and then 
the distances between query examples and each prototype followed by softmax, as in Eq.~(\ref{eq:proto}).

However, due to the sparsity and sporadicity (\ie, non-uniformly distributed) of the support examples,
the quality of prototypes varies substantially from episode to episode.
In order to better estimate class prototypes as well as better adapt to specific tasks,
we need to make full use of unlabeled query examples for semi-supervised or transductive inference.
As described in \cite{ren2018meta}, pseudo-labels obtained by Eq.~(\ref{eq:softmax_proto}) can be used to
update prototypes in a $K$-means step:
\begin{equation}
    \tilde{\mathbf{p}}_c = \frac{\sum\limits_{(x^s_i,y^s_i)\in\mathcal{D}^S}\mathbbm{1}_{[y^s_i=c]}f_{\phi^*}(x^s_i) + \sum\limits_{x^q_j\in\mathcal{D}^Q}p_c(x^q_j)f_{\phi^*}(x^q_j)}
                                {\sum\limits_{(x^s_i,y^s_i)\in\mathcal{D}^S}\mathbbm{1}_{[y^s_i=c]} + \sum\limits_{x^q_j\in\mathcal{D}^Q}p_c(x^q_j)}
\label{eq:kmeans}
\end{equation}

Another problem of centroid-nearest neighbor method is that, since each time only a few data points are sampled,
the data distribution of a single task drifts heavily from the overall data distribution.
Thus, certain transformations~\cite{wang2019simpleshot,hu2020leveraging} on the features are needed to calibrate them.
To this end, we propose CIPA that: 1) calibrates the features for better distance computation, and 
2) iteratively predicts pseudo-labels on unlabeled data and updates the estimation of prototypes progressively (Fig.~\ref{fig:overview} right panel).
The inference procedure is 
shown in Algorithm~\ref{algo:cipa}.

\begin{algorithm}
\label{algo:cipa}
\DontPrintSemicolon
    \KwInput{$\mathcal{T}^{(i)}_{novel}=(\mathcal{D}^S,\mathcal{D}^Q)$, $f_{\phi^*}$, $N_{iter}$, $\sigma$, $\tau$}
    \KwOutput{$\hat{p}(x^q_i)$ for each $x^q_i\in\mathcal{D}^Q$}
    \tcc{power transformation}
    $x^s_i \leftarrow \frac{(x^s_i)^\beta}{\|(x^s_i)^\beta\|}$,\
    $x^q_i \leftarrow \frac{(x^q_i)^\beta}{\|(x^q_i)^\beta\|}$ \\
    \tcc{zero-mean}
    $x^s_i \leftarrow x^s_i-\frac{1}{NK}\sum_{x^s_i\in\mathcal{D}^S}x^s_i$ \\
    $x^q_i \leftarrow x^q_i-\frac{1}{NQ}\sum_{x^q_i\in\mathcal{D}^Q}x^q_i$ \\
    \tcc{$l_2$ normalization}
    $x^s_i \leftarrow \frac{x^s_i}{\|x^s_i\|}$,\
    $x^q_i \leftarrow \frac{x^q_i}{\|x^q_i\|}$ \\
    \tcc{compute and update prototypes}
    Initialize $\mathbf{p}^{(0)}$ using Eq.~(\ref{eq:proto}).\\
    \For{$t=1,2,\dots N_{iter}$}
    {   
        $p^{(t)}(x^q)\leftarrow\mathsf{softmax}\big(\tau\cdot cos(x^q,\mathbf{p}^{(t-1)})\big)$ \\
        Compute new $\tilde{\mathbf{p}}^{(t)}$ using Eq.~(\ref{eq:kmeans}) and $p^{(t)}(x^q)$\\
        $\mathbf{p}^{(t)}\leftarrow \sigma\tilde{\mathbf{p}}^{(t)} + (1-\sigma)\mathbf{p}^{(t-1)}$
    }
    \tcc{predict using the final prototypes}
    $\hat{p}(x^q)\leftarrow\mathsf{softmax}\big(\tau\cdot cos(x^q,\mathbf{p}^{(N_{iter})})\big)$ \\
    \Return $\hat{p}(x^q)$
\caption{Calibrated Iterative Prototype Adaptation (CIPA)}
\end{algorithm}

In our experiments, we have found that this straightforward iterative inference algorithm can greatly improve FSL performance when unlabeled data is available.
Hu~\etal~\cite{hu2020leveraging} also uses an iterative approach to update class centers.
However, they assume that the test set has an equal number of examples for each class and use Sinkhorn mapping~\cite{cuturi2013sinkhorn} to find the best match.
While improved FSL performace is demonstrated, this is, to certain degree, due to the fact that episodes constructed under the evaluation protocols of FSL datasets do have uniform class distribution. Their method, therefore, may find it difficult in dealing with imbalanced classes.
Our CIPA does not rely on such assumptions and will work properly under class imbalance, which is critical for real-world applications.

In Algorithm~\ref{algo:cipa}, we use query examples to estimate pseudo-labels and update the prototypes.
However, CIPA is not limited to such a transductive setting and can be extended to semi-supervised FSL,
where another auxiliary set of unlabeled data is used instead of query examples themselves. We have conducted experiments and verified the effectiveness of CIPA for semi-supervised FSL (see the supplementary materials).

\section{Experiments}
\begin{table*}[t]
\small
\begin{center}
\begin{tabular}{l|l|c|cc|cc}
\hline
\multirow{2}{*}{\bf Setting} &\multirow{2}{*}{\bf Method} &\multirow{2}{*}{\bf Backbone} 
&\multicolumn{2}{c|}{\bf 5-way \emph{mini}-ImageNet} &\multicolumn{2}{c}{\bf 5-way \emph{tiered}-ImageNet} \\
& & &1-shot &5-shot &1-shot &5-shot \\
\hline
\multirow{9}{*}{In.} &TADAM~\cite{oreshkin2018tadam}          & ResNet12  &$58.50\pm0.30$ &$76.70\pm0.30$ &$-$            &$-$ \\
&ProtoNet~\cite{snell2017prototypical}\textsuperscript{$\dagger$}    & ResNet12      &$59.25\pm0.64$ &$75.60\pm0.48$ &$61.74\pm0.77$ &$80.00\pm0.55$ \\
&MetaOptNet-SVM~\cite{lee2019meta}       & ResNet12  &$62.64\pm0.61$ &$78.63\pm0.46$ &$65.99\pm0.72$ &$81.56\pm0.53$ \\
&SNAIL~\cite{mishra2017simple}           & ResNet15  &$55.71\pm0.99$ &$68.88\pm0.92$ &$-$            &$-$ \\
&SimpleShot~\cite{wang2019simpleshot}    & ResNet18  &$62.85\pm0.20$ &$80.02\pm0.14$ &$69.09\pm0.22$ &$84.58\pm0.16$ \\
&DeepEMD~\cite{zhang2020deepemd}         & ResNet12  &$65.91\pm0.82$ &$82.41\pm0.56$ &$71.16\pm0.87$ &$86.03\pm0.58$ \\
&LEO~\cite{rusu2018meta}                 & WRN-28-10 &$61.76\pm0.08$ &$77.59\pm0.12$ &$66.33\pm0.05$ &$81.44\pm0.09$ \\
&CC+rot~\cite{gidaris2019boosting}       & WRN-28-10 &$62.93\pm0.45$ &$79.87\pm0.33$ &$70.53\pm0.51$ &$84.98\pm0.36$ \\
&S2M2\textsubscript{R}~\cite{mangla2020charting}
                                         & WRN-28-10  &$64.93\pm0.18$ &$83.18\pm0.11$ &$73.71\pm0.22$ &\bftab88.59 $\pm$ 0.14 \\
\hline
\multirow{7}{*}{Trans.} &TPN~\cite{liu2018learning}    & ResNet12  &$59.46$    &$75.65$    &$58.68^\S$   &$74.26^\S$ \\
&Trans. Fine-Tuning~\cite{dhillon2019baseline} & ResNet12  &$62.35\pm0.66$ &$74.53\pm0.54$ &$68.41\pm0.73$ &$83.41\pm0.52$ \\
&TEAM~\cite{qiao2019transductive}        & ResNet18  &$60.07$    &$75.90$    &$-$    &$-$ \\
&LR + ICI~\cite{wang2020instance}          & ResNet12  &$66.80$    &$79.26$    &$80.79$    &\bftab87.92 \\
&FEAT~\cite{ye2020few}                   & ResNet18  &$66.78\pm0.20$ &$82.05\pm0.14$ &$70.80\pm0.23$ &$84.79\pm0.16$ \\
&EPNet~\cite{rodriguez2020epnet}      & ResNet12  &66.50 $\pm$ 0..89  &81.06 $\pm$ 0.60  &76.53 $\pm$ 0.87  &\bftab87.32 $\pm$ 0.64 \\
&LaplacianShot~\cite{ziko2020laplacian}  & ResNet18  &$72.11\pm0.19$ &$82.31\pm0.14$ &$78.98\pm0.21$ &$86.39\pm0.16$ \\
&HCT\textsubscript{R} + CIPA (ours)      & ResNet12  &\bftab76.94 $\pm$ 0.24  &\bftab85.10 $\pm$ 0.14 &\bftab81.70 $\pm$ 0.25  &\bftab87.91 $\pm$ 0.15 \\
\hline
\end{tabular}
\end{center}
\vspace{-0.5em}
\caption{Results on \emph{mini}-ImageNet and \emph{tiered}-ImageNet.
In. and Trans. stand for inductive and transductive, respectively.
Methods marked with $\dagger$ are reported in Lee~\etal~\cite{lee2019meta},
while those with \S\ are from Wang~\etal~\cite{wang2020instance}.
Our accuracies are averaged over 10k episodes.}
\label{tab:imagenet}
\vspace{-1em}
\end{table*}

\subsection{Settings}
\label{sec:settings}
\noindent{\bf Datasets.} We conducted experiments on five FSL datasets:
1) {\bf \emph{mini}-ImageNet}~\cite{vinyals2016matching} is derived from the ILSVRC2012~\cite{russakovsky2015imagenet} dataset.
It contains 100 randomly sampled classes and is split into 64, 16 and 20 classes for
train, validation and test, respectively. Each class has 600 images, which are resized into $84\times 84$.
2) {\bf \emph{tiered}-ImageNet}~\cite{ren2018meta} is also a subset of ILSVRC2012~\cite{russakovsky2015imagenet}.
It contains in total 34 super categories and is split into 20, 6 and 8 for train, validation and test, respectively.
The corresponding class numbers are 351, 97 and 160. On average, each class has around 1280 images.
Similar to \emph{mini}-ImageNet, all images are resized into $84\times 84$.
3) {\bf CIFAR\_FS}~\cite{bertinetto2018meta} is a few-shot learning dataset that contains all 100 classes from CIFAR100~\cite{krizhevsky2009learning}.
The dataset is randomly split into 64, 16 and 20 classes for train, validation and test. Each class has 600 images of size $32\times 32$.
4) {\bf FC\_100}~\cite{bertinetto2018meta} is also derived from CIFAR100~\cite{krizhevsky2009learning}.
But it is instead split into 60, 20 and 20 classes that are from 12, 4 and 4 super categories.
It is like the ``tiered'' version of CIFAR\_FS. Similarly, it has 600 images of size $32\times 32$ for each class.
5) {\bf CUB}~\cite{wah2011caltech} is a dataset of 200 fine-grained bird species. 
We follow~\cite{chen2019closer} to split the dataset into 100, 50 and 50 for train, validation and test.
This dataset only has around 59 images for each class.
For all these five datasets, we resize the images into $84\times 84$ if they are not so already.

\vspace{1mm}
\noindent{\bf Training settings.} In all our experiments, we use ResNet-12~\cite{chen2020new} as our backbone network.
Every ResNet block contains three $3\times3$ convolutional layers, each followed by a BatchNorm layer and ReLU activation. 
There are four blocks and the number of filters for each block increases as 64, 128, 256 and 512.
There is a final average pooling layer that shrinks the output tensor into a 512-dim vector.
To train the backbone ResNet-12, we use the Adam optimizer with a learning rate of 0.001 and train for 300 epochs (60 on \emph{tiered}-ImageNet).
During the first $1/3$ of total epochs we use $\mathcal{L}_{ce}+\mathcal{L}_{rot}$,
for the remaining $2/3$ of the epochs we add the $\mathcal{L}_{hct}$ loss term.
To interpolate examples, we use $\alpha=2$ to sample $\lambda{\sim}Beta(\alpha,\alpha)$ by default unless stated otherwise.
For the weak augmentation, we use random crop and random flip at 50\% chance.
For the strong augmentation, we follow FixMatch~\cite{sohn2020fixmatch} and use RandAugment~\cite{cubuk2020randaugment}.
Each time, 2 out of 14 augmentations are randomly selected and applied to the image,
after which a random square region in the image is cut out \cite{devries2017improved}.
We use the same settings for all datasets to obtain our main results.
Performance on validation data is monitored during training for model selection.

\vspace{1mm}
\noindent{\bf Evaluation settings.} In the test phase, we fix the trained backbone network and use it as a feature extractor.
The extracted features of the support and query samples are used by CIPA to predict their classes. We use $\beta=0.5$, $\sigma=0.2$ and $N_{iter}=20$ for all experiments.
In each experiment, a number of novel episodes (600 or 10,000) are sampled.
Each episode contains $N$ classes, and each class has $K$ support and 15 query examples.
We report the average accuracy and 95\% confidence interval as performance measurements.

\subsection{Main results}
\begin{table*}[t]
\small
\begin{center}
\begin{tabular}{l|l|c|cc|cc}
\hline
\multirow{2}{*}{\bf Setting} &\multirow{2}{*}{\bf Method} &\multirow{2}{*}{\bf Backbone} &\multicolumn{2}{c|}{\bf 5-way CIFAR\_FS} &\multicolumn{2}{c}{\bf 5-way FC100} \\
& & &1-shot &5-shot &1-shot &5-shot \\
\hline
\multirow{7}{*}{In.} &ProtoNet~\cite{snell2017prototypical}\textsuperscript{$\dagger$}
                                         & ResNet12  &$72.2\pm0.7$  &$83.5\pm0.5$  &$37.5\pm0.6$  &$52.5\pm0.6$ \\
&MetaOptNet-SVM~\cite{lee2019meta}       & ResNet12  &$72.0\pm0.7$  &$84.2\pm0.5$  &$41.1\pm0.6$  &$55.5\pm0.6$ \\
&TADAM~\cite{oreshkin2018tadam}          & ResNet12  &$-$    &$-$    &$40.1\pm0.4$   &$56.1\pm0.4$ \\
&SimpleShot~\cite{wang2019simpleshot}    & ResNet10  &$-$    &$-$    &$40.13\pm0.18$ &$53.63\pm0.18$ \\
&DeepEMD~\cite{zhang2020deepemd}         & ResNet12  &$-$    &$-$    &$46.47\pm0.78$ &$63.22\pm0.71$ \\
&CC+rot~\cite{gidaris2019boosting}       & WRN-28-10 &$76.09\pm0.30$ &$87.83\pm0.21$ &$-$ &$-$ \\
&S2M2\textsubscript{R}~\cite{mangla2020charting}       & WRN-28-10 &$74.81\pm0.19$ &$87.47\pm0.13$ &$-$ &$-$ \\
\hline
\multirow{6}{*}{Trans.} &TPN~\cite{liu2018learning}              & ResNet12  &$65.89^\S$    &$79.38^\S$  &$-$    &$-$ \\
&TEAM~\cite{qiao2019transductive}        & ResNet18  &$70.43$    &$81.25$    &$-$    &$-$ \\
&Transductive Fine-Tuning~\cite{dhillon2019baseline} & ResNet12  &$70.76\pm0.74$ &$81.56\pm0.53$ &$41.89\pm0.59$ &$54.96\pm0.55$ \\
&LR + ICI~\cite{wang2020instance}          & ResNet12  &$73.97$    &$84.13$    &$-$    &$-$ \\
&HCT\textsubscript{R} + CIPA (ours)      & ResNet12  &\bftab85.72 $\pm$ 0.21  &\bftab89.69 $\pm$ 0.14  &\bftab53.30 $\pm$ 0.25  &\bftab64.90 $\pm$ 0.20 \\
\hline
\end{tabular}
\end{center}
\vspace{-0.5em}
\caption{Results on CIFAR\_FS and FC100.
Our accuracies are averaged over 10k episodes.}
\label{tab:cifar}
\vspace{-0.5em}
\end{table*}

\noindent{\bf Standard few-shot learning.} 
We separate comparison methods into the inductive and transductive groups. In the transductive group, we choose published SOTA algorithms with a similar backbone (\ie, ResNet12, ResNet 18) while relaxing such requirement in the inductive group for fair comparison. 

We summarize the results on \emph{mini}-ImageNet and \emph{tiered}-ImageNet in Table~\ref{tab:imagenet}. We can see that our method, HCT\textsubscript{R}+ CIPA, has achieved the best performance
among comparison methods on the \emph{mini}-ImageNet dataset.
Comparing to LaplacianShot~\cite{ziko2020laplacian}, the best performing method reported using a ResNet18 backbone, we achieve more than $4\%$ and nearly $3\%$ improvements on 1-shot and 5-shot, respectively.
As for \emph{tiered}-ImageNet, HCT\textsubscript{R}+CIPA yields the best performance on 1-shot while
being on par with S2M2\textsubscript{R}~\cite{mangla2020charting} and LR+ICI~\cite{wang2020instance} on 5-shot.
The results on CIFAR\_FS and FC100 are summarized in Table~\ref{tab:cifar}. Similarly, our method achieves the best performance across all settings. Note that some of the methods in the inductive group, such as CC+rot and S2M2\textsubscript{R}, use a larger network (\eg, WRN-28-10). Our method still outperforms them, showing that our training method combined with transductive inference can compensate for the disadvantages of using a lighter network.

CUB is a different dataset from the previous ones in that it contains fine-grained bird species as classes rather than generic objects.
We summarize results on CUB in Table~\ref{tab:cub}.
Again, our approach has achieved the best performance on both 1-shot and 5-shot with an improvement of ${\sim}5\%$ and ${\sim}2\%$, respectively, over LR+ICI~\cite{wang2020instance}, the best reported method in literature using a ResNet12 backbone. Notably, even comparing to transductive methods with a larger backbone of WRN-28-10, \eg, PT+MAP~\cite{hu2020leveraging} that achieves $91.55\pm0.19$ and $93.99\pm0.10$ for 1- and 5-shot on CUB, respectively, our method still remains the best. The results on CUB strongly suggest that regularizing learned embedding in a wider extent and across network layers can help to learn rich and robust representations to significantly benefit FSL on fine-grained classes. 


\begin{table}[t]
\small
\setlength{\tabcolsep}{2pt}
\begin{center}
\begin{tabular}{l|c|c c}
\hline
\multirow{2}{*}{\bf Method} &\multirow{2}{*}{\bf Backbone} &\multicolumn{2}{c}{\bf 5-way CUB}\\
 & &1-shot &5-shot \\
\hline
DeepEMD~\cite{zhang2020deepemd}         & ResNet12  &$75.65\pm0.83$ &$88.69\pm0.50$ \\
S2M2\textsubscript{R}~\cite{mangla2020charting}       & WRN-28-10 &$80.68\pm0.81$ &$90.85\pm0.44$ \\
\hline
TEAM~\cite{qiao2019transductive}        & ResNet18  &$80.16$    &$87.17$ \\
LaplacianShot~\cite{ziko2020laplacian}  & ResNet18  &$80.96$    &$88.68$ \\
LR+ICI~\cite{wang2020instance}          & ResNet12  &$88.06$    &$92.53$ \\
HCT\textsubscript{R} + CIPA (ours)      & ResNet12  &\bftab93.03 $\pm$ 0.15  &\bftab94.90 $\pm$ 0.08 \\
\hline
\end{tabular}
\end{center}
\vspace{-0.5em}
\caption{Results on CUB. Our accuracies are averaged over 10k episodes.}
\label{tab:cub}
\vspace{-1em}
\end{table}

\vspace{1mm}
\noindent{\bf Cross-domain FSL.}
\begin{table}[t]
\small
\setlength{\tabcolsep}{2pt}
\begin{center}
\begin{tabular}{l|c|c c}
\hline
\multirow{2}{*}{\bf Method} &\multirow{2}{*}{\bf Backbone} &\multicolumn{2}{c}{\bf\emph{mini}-ImageNet$\rightarrow$ CUB}\\
 & &1-shot &5-shot \\
\hline
Mat. Net~\cite{vinyals2016matching} + FT\textsuperscript{\dagger}         & ResNet10  &$36.61\pm0.53$ &$55.23\pm0.83$ \\
Rel. Net~\cite{sung2018learning} + FT\textsuperscript{\dagger}         & ResNet10  &$44.07\pm0.77$ &$59.46\pm0.71$ \\
S2M2\textsubscript{R}~\cite{mangla2020charting}       & WRN-28-10 &$48.24\pm0.84$ &$70.44\pm0.75$ \\
\hline
GNN~\cite{garcia2017few} + FT\textsuperscript{\dagger}         & ResNet10  &$47.47\pm0.75$ &$66.98\pm0.68$ \\
LaplacianShot~\cite{ziko2020laplacian}  & ResNet18  &$55.46$    &$66.33$ \\
HCT\textsubscript{R} + CIPA (ours)      & ResNet12  &\bftab62.15 $\pm$ 1.08  &\bftab74.25 $\pm$ 0.77 \\
\hline
\end{tabular}
\end{center}
\vspace{-0.5em}
\caption{Results for cross-domain FSL. Our accuracies are averaged over 600 episodes.
\dagger\ are reported in Tseng~\etal~\cite{tseng2020cross}.}
\label{tab:cross}
\vspace{-1em}
\end{table}
To study the robustness of representations learned via our HCT across datasets with certain amounts of covariate shift, we evaluate its performance under cross-domain scenarios as an outreaching test. 
We train models on \emph{mini}-ImageNet and test them on CUB.
From Table~\ref{tab:cross},
our method, HCT\textsubscript{R}+CIPA, achieves the best performance on both 1-shot and 5-shot tasks, with an improvement of $7\%$ and $8\%$ over LaplacianShot, respectively. This manifests that our method not only works under in-domain settings,
but also can generalize well under the more challenging cross-domain settings.

\subsection{Ablation studies}
\noindent{\bf HCT for embedding learning.}
\begin{table*}[h!]
\small
\begin{center}
\begin{tabular}{l|cccc|cc|cc|cc}
\hline
\multirow{2}{*}{\bf Method} &\multicolumn{4}{c|}{\bf Train} &\multicolumn{2}{c}{\bf PN} &\multicolumn{2}{c}{\bf SemiPN} &\multicolumn{2}{c}{\bf CIPA}\\
& $\mathcal{L}_{ce}$ & $\mathcal{L}_{mm}$ & $\mathcal{L}_{hct}$ & $\mathcal{L}_{rot}$ &1-shot &5-shot &1-shot &5-shot &1-shot &5-shot\\
\hline
\multicolumn{11}{c}{\bf \emph{mini}-ImageNet} \\
\hline
Classifier Baseline & $\checkmark$ & & & &$56.48$ &$75.62$ &$66.14$ &$77.72$ &$70.84$ &$80.59$ \\
Manifold Mixup  & $\checkmark$ & $\checkmark$ &  & &$57.07$ &$78.09$ &$68.25$ &$80.12$ &$73.69$ &$83.06$ \\
HCT & $\checkmark$ &  & $\checkmark$ & &$58.54$ &$78.43$ &$69.38$ &$80.33$ &$74.74$ &$82.91$ \\
S2M2\textsubscript{R}  & $\checkmark$ & $\checkmark$ &  & $\checkmark$ &$59.66$ &$77.60$ &$68.77$ &$79.99$ &$76.54$ &$85.16$ \\
HCT\textsubscript{R}  & $\checkmark$ &  & $\checkmark$ & $\checkmark$ &$60.33$ &$77.66$ &$69.38$ &$80.32$ &$77.26$ &$84.89$ \\
\hline
\multicolumn{11}{c}{\bf CUB} \\
\hline
Classifier Baseline & $\checkmark$ & & & &$67.56$ &$85.63$ &$79.57$ &$87.95$ &$84.96$ &$89.84$ \\
Manifold Mixup  & $\checkmark$ & $\checkmark$ &  & &$65.78$ &$86.53$ &$79.26$ &$88.84$ &$86.12$ &$90.95$ \\
HCT  & $\checkmark$ &  & $\checkmark$ & &$67.90$ &$86.73$ &$80.43$ &$89.20$ &$86.79$ &$91.02$ \\
S2M2\textsubscript{R}  & $\checkmark$ & $\checkmark$ &  & $\checkmark$ &$73.84$ &$88.26$ &$83.52$ &$90.05$ &$88.40$ &$91.93$ \\
HCT\textsubscript{R}  & $\checkmark$ &  & $\checkmark$ & $\checkmark$ &$81.68$ &$92.39$ &$89.47$ &$93.22$ &$93.27$ &$94.77$ \\
\hline
\end{tabular}
\end{center}
\vspace{-0.5em}
\caption{Ablation study on HCT. Accuracies are averaged over 600 episodes.}
\label{tab:hct}
\vspace{-1.5em}
\end{table*}
\begin{table}[t]
\small
\setlength{\tabcolsep}{4pt}
\begin{center}
\begin{tabular}{cccc|cc|cc}

\hline
\multicolumn{8}{c}{\bf \emph{mini}-ImageNet }  \\
& center & $l_2$ norm. & pow. & $\sigma$ & $N_{iter}$ & 1-shot& 5-shot\\
\hline
(a) &  &  &  & N/A & 0 & 60.33 & 77.66 \\
(b) &$\checkmark$ &  &  & N/A & 0 & 63.48 & 78.39 \\
(c) &$\checkmark$ & $\checkmark$ &   & N/A & 0 & 63.48 & 78.73 \\
(d) &$\checkmark$ & $\checkmark$ & $\checkmark$ & N/A & 0 & 65.96 & 81.35 \\
\hline
(e) &$\checkmark$ & $\checkmark$ & $\checkmark$ & 1.0 & 1 & 72.92 & 83.94 \\
(f) &$\checkmark$ & $\checkmark$ & $\checkmark$ & 1.0 & 20 & \textbf{78.19} & 84.74 \\
(g) &$\checkmark$ & $\checkmark$ & $\checkmark$ & 0.2 & 20 & 77.26 & \textbf{84.89} \\
\hline
\end{tabular}
\end{center}
\vspace{-0.5em}
\caption{Ablation study on CIPA.
Accuracies are averaged over 600 episodes on \emph{mini}-ImageNet.}
\label{tab:cipa}
\vspace{-2em}
\end{table}
To better understand how each component of HCT affects the learning of representations,
we design our experiments in two directions (Table~\ref{tab:hct}):
1) how the embedding is trained (row-wise) and
2) what inference algorithm is used (column-wise).
To train an embedding model, we start with ``Classifier Baseline'',
which only uses $\mathcal{L}_{ce}$.
We then add $\mathcal{L}_{mm}$ for Manifold Mixup,
or $\mathcal{L}_{hct}$ for our HCT.
Beyond these, adding another rotation loss $\mathcal{L}_{rot}$ yields S2M2\textsubscript{R} and HCT\textsubscript{R}.
As for inference, we compare our CIPA against ProtoNet~\cite{snell2017prototypical}, a centroid-nearest neighbor based method, and SemiPN~\cite{ren2018meta}, an extension of ProtoNet that makes use of unlabeled data. 

From Table~\ref{tab:hct}, we have several observations:
1) Comparing three inference algorithms, our CIPA is consistently the best across all experiments.
2) Comparing Classifier Baseline and HCT, adding $\mathcal{L}_{hct}$ leads to $2{\sim}3\%$ improvements across all inference algorithms on \emph{mini}-ImageNet. On CUB, the improvements are marginal in inductive settings (\ie, using PN), but more noticeable in transductive settings (\ie, using SemiPN or CIPA).
3) Comparing S2M2\textsubscript{R} and HCT\textsubscript{R}, \ie, $\mathcal{L}_{mm}$ v.s. $\mathcal{L}_{hct}$ on top of $\mathcal{L}_{ce}+\mathcal{L}_{rot}$,
the performances are at the same level on \emph{mini}-ImageNet. On the CUB dataset, we see a significant improvement from our HCT.
For a more thorough comparison between Manifold Mixup and HCT on different $\alpha$ values, please see details in the supplementary materials.

Based on these observations, we conclude that the benefit introduced by HCT depends on two factors:
1) the method onto which HCT is added and 2) the dataset to which HCT is applied.
Overall, HCT promises improvements for FSL, especially when used in combination with our CIPA.

\vspace{1mm}
\noindent{\bf CIPA for transductive inference.}
We then study how each component of our iterative algorithm CIPA affects the final performance in Table~\ref{tab:cipa}.
Comparing rows (a) and (b),
we find that simply subtracting the mean induces
a nearly ${\sim}3\%$ improvement on 1-shot.
This indicates the existence of shift between the data distributions of few-shot tasks and the true data distribution
and a simple centering can effectively compensate for such a shift, especially for 1-shot.
Comparing rows (c) and (d), we note that power transform~\cite{hu2020leveraging} also introduces an improvement of ${\sim}2\%$ for both 1- and 5-shot. 
As expected, the greatest increase, $7\%$ on 1-shot and ${\sim}3\%$ on 5-shot, comes from adapting the prototypes using unlabeled examples (row (e) vs. (d)). 
Tuning the adaptation parameters also helps improving the performance of CIPA (see rows (e) to (g)).
\begin{figure}[t]
\begin{center}
   \includegraphics[width=0.85\linewidth]{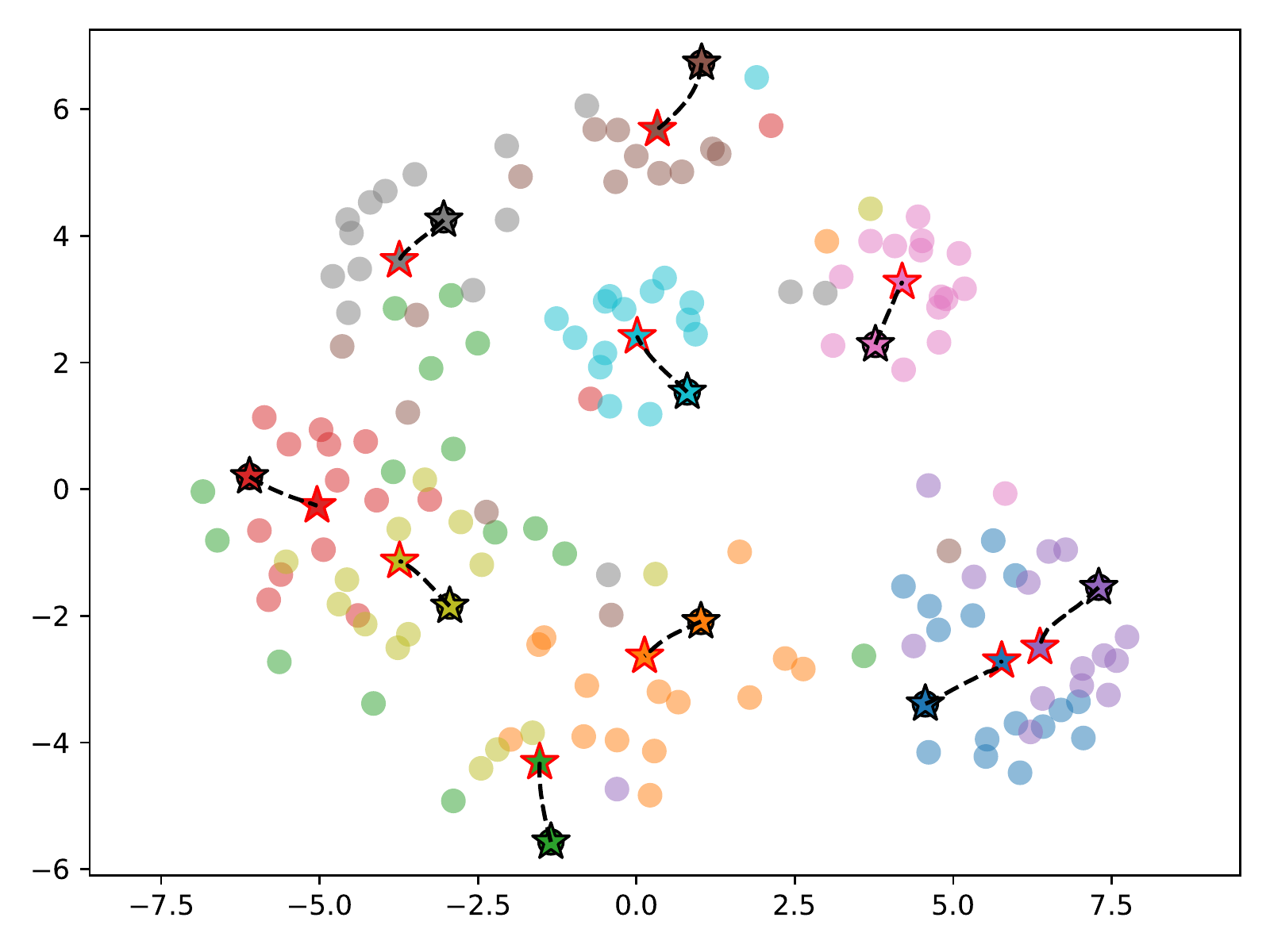}
\end{center}
\vspace{-1em}
   \caption{Visualization of a 10-way 1-shot task from \emph{mini}-ImageNet using t-SNE~\cite{maaten2008visualizing}.
   Colors represent ground truth labels.
   ``$\medwhitestar$'' with black outline are initial prototypes and those with red outline are adapted prototypes. Dashed lines show how they gradually evolve.
   Circles with no outline are unlabeled query examples and those with black outline are labeled support examples.
   Note that there is only one labeled example per class, so the initial prototypes overlap with them
   (zoom in for a better view).}
\label{fig:iter_cos}
\vspace{-1em}
\end{figure}
Finally 
we plot a randomly sampled task in Fig.~\ref{fig:iter_cos} for an intuitive understanding of the inner workings and effect of our CIPA.
We can see that, at the beginning, prototypes are just those 1-shot labeled examples,
which are sub-optimal for prediction. After the iterative adaptation procedure,
estimated prototypes gradually move toward the center of each class, generating better predictions.

\vspace{-0.5em}
\section{Conclusion}
\vspace{-0.5em}
Few-Shot Learning is a critical problem to be addressed for a wider utilization of deep learning but still remains challenging.
In this paper, we tackled two longstanding difficulties in FSL.
1) To generalize from base to novel classes with semantic gaps, we proposed hybrid consistency training, \ie, HCT, a combination of interpolation consistency and data augmentation consistency to regularize the learning of representations.
2) To bridge the gap between sparse support and query examples, we developed a transductive inference algorithm, \ie, CIPA, to calibrate features and use unlabeled
data to adapt prototypes iteratively. Through extensive experiments, we have shown that our method can achieve SOTA performance on all five FSL datasets.
Ablation studies also justified the necessity and quantified the effectiveness of each component in HCT and CIPA.

\section{Acknowledgments}
This material is based upon work supported by the United States Air Force under Contract No. FA850-19-C-0511.  Any opinions, findings and conclusions or recommendations expressed in this material are those of the author(s) and do not necessarily reflect the views of the United States Air Force. 

{\small
\bibliographystyle{ieee_fullname}
\bibliography{egbib}
}

\clearpage
\appendix
\section{Manifold Mixup vs. HCT}
To have a thorough comparison between Manifold Mixup~\cite{verma2019manifold} and our proposed Hybrid Consistency Training (HCT), we train models using these two approaches with different $\alpha$ values. Note that $\alpha$ determines the distribution from which the weight $\lambda$ that balances the linear combination of the two samples is drawn: $\lambda{\sim}Beta(\alpha,\alpha)$. We keep all other hyper-parameters exactly the same so that
the changes in accuracies are only caused by the different behaviors between Manifold Mixup and HCT, \ie, $\mathcal{L}_{mm}$ v.s. $\mathcal{L}_{hct}$.
Table \ref{tab:apdx_hct} shows the results on both the \emph{mini}-ImageNet and CUB datasets.
We can observe that, overall, HCT achieves better performance than Manifold Mixup.
The improvement is more obvious on 1-shot tasks, while less noticeable on 5-shot tasks.
This is reasonable since performance differences among Few-Shot Learning (FSL) methods tend to decrease as more labeled examples are used. These results prove that our proposed HCT is a better alternative than Manifold Mixup on FSL problems.

\begin{table*}[t]
\small
\begin{center}
\begin{tabular}{l|cccc|cc|cc|cc}
\hline
\multirow{2}{*}{\bf Method} &\multicolumn{4}{c|}{\bf Train} &\multicolumn{2}{c}{\bf PN} &\multicolumn{2}{c}{\bf SemiPN} &\multicolumn{2}{c}{\bf CIPA}\\
& $\mathcal{L}_{ce}$ & $\mathcal{L}_{mm}$ & $\mathcal{L}_{hct}$ & $\mathcal{L}_{rot}$ &1-shot &5-shot &1-shot &5-shot &1-shot &5-shot\\
\hline
\multicolumn{11}{c}{\bf \emph{mini}-ImageNet} \\
\hline
Manifold Mixup ($\alpha=0.5$) & $\checkmark$ & $\checkmark$ &  & &$57.48$ &$77.04$ &$67.19$ &$78.96$ &$71.85$ &$81.32$ \\
Manifold Mixup ($\alpha=1.0$) & $\checkmark$ & $\checkmark$ &  & &$57.07$ &$78.09$ &$68.25$ &$80.12$ &$73.69$ &$83.06$ \\
Manifold Mixup ($\alpha=2.0$) & $\checkmark$ & $\checkmark$ &  & &$56.42$ &$77.81$ &$67.54$ &$79.96$ &$73.71$ &$82.69$ \\
\hline
HCT ($\alpha=0.5$)& $\checkmark$ &  & $\checkmark$ & &$58.47$ &$78.53$ &$69.00$ &$80.31$ &$74.69$ &$83.10$ \\
HCT ($\alpha=1.0$)& $\checkmark$ &  & $\checkmark$ & &$58.54$ &$78.43$ &$69.38$ &$80.33$ &$74.74$ &$82.91$ \\
HCT ($\alpha=2.0$)& $\checkmark$ &  & $\checkmark$ & &$57.38$ &$78.54$ &$68.31$ &$80.69$ &$74.09$ &$83.26$ \\
\hline
\multicolumn{11}{c}{\bf CUB} \\
\hline
Manifold Mixup ($\alpha=0.5$) & $\checkmark$ & $\checkmark$ &  & &$66.32$ &$86.57$ &$79.82$ &$88.94$ &$86.26$ &$90.95$ \\
Manifold Mixup ($\alpha=1.0$) & $\checkmark$ & $\checkmark$ &  & &$65.78$ &$86.53$ &$79.26$ &$88.84$ &$86.12$ &$90.95$ \\
Manifold Mixup ($\alpha=2.0$) & $\checkmark$ & $\checkmark$ &  & &$66.28$ &$86.63$ &$79.82$ &$89.12$ &$86.91$ &$91.11$ \\
\hline
HCT ($\alpha=0.5$)& $\checkmark$ &  & $\checkmark$ & &$68.97$ &$86.80$ &$80.53$ &$88.99$ &$86.19$ &$90.73$ \\
HCT ($\alpha=1.0$)& $\checkmark$ &  & $\checkmark$ & &$67.90$ &$86.73$ &$80.43$ &$89.20$ &$86.79$ &$91.02$ \\
HCT ($\alpha=2.0$)& $\checkmark$ &  & $\checkmark$ & &$67.67$ &$86.89$ &$80.40$ &$89.20$ &$87.34$ &$91.11$ \\
\hline
\end{tabular}
\end{center}
\vspace{-0.5em}
\caption{Comparison between Manifold Mixup and HCT on various $\alpha$ values. Accuracies are averaged over 600 episodes.}
\label{tab:apdx_hct}
\vspace{-1em}
\end{table*}

\section{Semi-supervised FSL}
Our proposed Calibrated Iterative Prototype Adaptation (CIPA) algorithm can not only be used for transductive inference,
but also be naturally extended to the semi-supervised FSL setting, which is first proposed in SemiPN~\cite{ren2018meta}.
The difference between semi-supervised FSL and transductive FSL is that
the former uses a separate auxiliary set of unlabeled examples to improve performance on query examples,
while the latter uses query examples themselves for this purpose.

For semi-supervised FSL, we split the novel data into labeled and unlabeled sets (e.g., 60\% as labeled and 40\% as unlabeled).
When generating test episodes, we always sample support and query examples from the labeled split (e.g., th 60\% split),
and sample auxiliary examples from the unlabeled split.
When updating the prototypes, only auxiliary examples are used.
After the class prototypes have been updated, they are used for prediction on the query examples.
Since semi-supervised FSL is not transductive, statistics of query examples should not be used.
Thus, for query examples,
we remove the zero-mean transformation and only perform the power transformation and $l_2$ normalization. In each episode, we use one support example, $M=1,2,4,\dots,128$ unlabeled examples, and 15 query examples per class.

\begin{figure*}[h]
\begin{center}
   \includegraphics[width=0.65\linewidth]{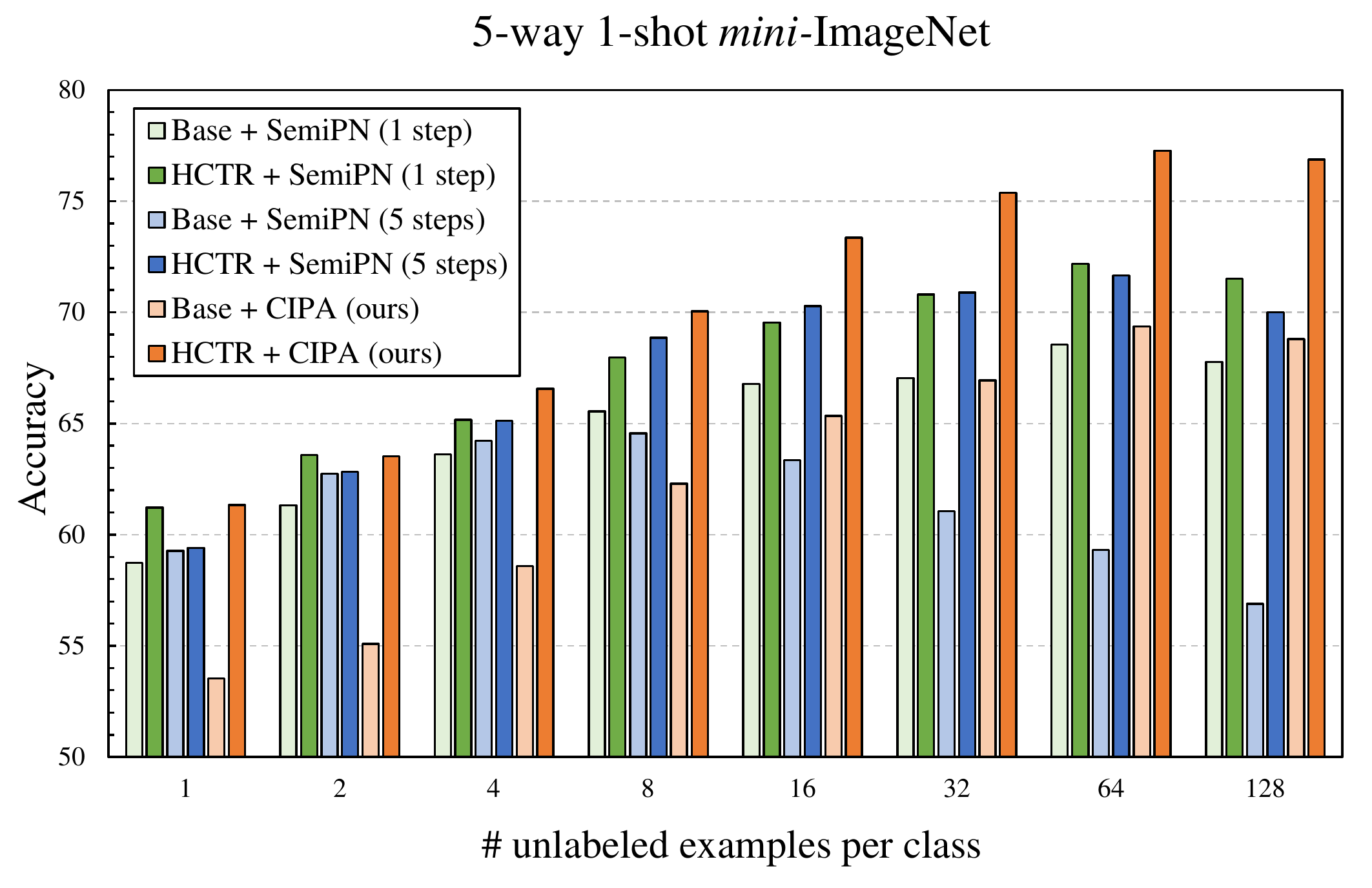}
\end{center}
\vspace{-1em}
   \caption{Bar chart of semi-supervised FSL results.
   Accuracies are averaged over 600 episodes. We omit the confidence intervals for clearer view. }
\label{fig:apdx_ssfsl}
\end{figure*}

The 5-way 1-shot results on \emph{mini}-ImageNet are shown in Figure~\ref{fig:apdx_ssfsl}.
Generally, as more unlabeled examples are used, the performance increases and saturates at certain $M$. For instance, the performance of HCT\textsubscript{R}+SemiPN saturates at $M=16$ whereas the performance of HCT\textsubscript{R}+CIPA saturates at a later point with $M=64$. This leaves a large room for CIPA to achieve higher performance, as indicated by an increasing lead of HCT\textsubscript{R}+CIPA as $M$ increases. 


Overall, we observe that HCT\textsubscript{R} and CIPA consistently outperform their counterpart: Classifier Baseline (denoted as Base in the chart to save space) and SemiPN, respectively. It is, therefore, expected that combining HCT\textsubscript{R} and CIPA yields consistently superior performance among all methods. However, it is worth noting the interesting behaviors of the weaker combinations such as Base+CIPA and HCT\textsubscript{R}+SemiPN as $M$ increases. Indeed, with more unlabeled data the strength of HCT\textsubscript{R} and CIPA starts to merge and eventually can compensate for the previously worse performance. We show two examples below. 

Comparing Base+SemiPN(1 step) and Base+SemiPN(5 steps),
we note that the latter one has a better performance for a smaller $M$ while its performance saturate quickly (at $M=4$) and starts degrading.
The reason of this unexpected trend might be that, when there is more unlabeled data, the prototypes are easier to be distracted by noisy pseudo-labels in more iterations.
However, when a better embedding model e.g., HCT\textsubscript{R} is used, this trend can be fixed to certain degree. Comparing HCT\textsubscript{R}+SemiPN(1 step) and HCT\textsubscript{R}+ SemiPN(5 steps), the turning point is at $M=64$. This demonstrates the robustness of the embedding leaned by HCT\textsubscript{R}.

Comparing Base+SemiPN(1 step) and Base+CIPA, we can see that, when only a few unlabeled examples are available, CIPA produces inferior results.
It catches up and achieves higher numbers as $M$ increases.
Our explanation is that, since CIPA calibrate the support data distribution and unlabeled data distribution separately,
when both of them are sparse, the calibration might not work properly.
This is also why it achieves better performance for larger $M$, where calibrating on more unlabeled data makes distance computation better. This verifies the strong adaptation capability of CIPA. 

To conclude, a better embedding model (i.e., HCT\textsubscript{R}) and 
a calibrated adaptive inference (i.e., CIPA) are both needed to achieve optimal FSL performance.

\end{document}